\documentclass{article}
\usepackage{amsfonts,amsmath,amsthm,amssymb}

\date{}

\usepackage{algorithm}
\usepackage{algorithmic}
\usepackage{subfigure}
\usepackage{epsfig}
\usepackage{graphicx}
\usepackage{url}
\usepackage{multirow}
\usepackage{amssymb}
\usepackage{mathrsfs}
\usepackage{epstopdf}
\usepackage{multicol}
\usepackage{amsmath}
\usepackage{mathtools}

\usepackage{multicol}
\usepackage{wrapfig}
\usepackage{wrapfig,lipsum,booktabs}

\usepackage{multirow}

\usepackage{pifont}% http://ctan.org/pkg/pifont
\newcommand{\cmark}{\ding{51}}%
\newcommand{\xmark}{\ding{55}}%

\allowdisplaybreaks[4]

\newtheorem{theorem}{\textbf{Theorem}}
\newtheorem{assumption}{\textbf{Assumption}}
\newtheorem{lemma}{\textbf{Lemma}}
\newtheorem{corollary}{\textbf{Corollary}}
\newtheorem{remark}{\textbf{Remark}}

\numberwithin{equation}{section}

\theoremstyle{plain}

\theoremstyle{definition}

\title{On the Convergence of Momentum-Based Algorithms for \\
	Federated Bilevel Optimization Problems}

\author{
	Hongchang Gao \thanks{Temple University, {\tt hongchang.gao@temple.edu}} 
	}
\begin{document}
\maketitle
%%-----------------------------
%%      the top matter
%%-----------------------------

%
\begin{abstract} 
In  this paper, we studied the federated bilevel optimization problem, which has widespread applications in machine learning.  In particular, we developed two momentum-based algorithms for optimizing this kind of problem, and  established the convergence rate of our two algorithms, providing the sample and communication complexities. Importantly, to the best of our knowledge, our convergence rate is the first one  achieving the linear speedup with respect to the number of devices for federated bilevel optimization algorithms.  At last, our extensive experimental results confirm the effectiveness of our two algorithms.
\end{abstract}

%

%%-----------------------------
%%      your text
%%-----------------------------
\section{Introduction}
In recent years, Federated Learning has attracted a surge of attention due to its great potential application in numerous real-world  machine learning tasks.  As such, a wide variety of federated optimization algorithms have been proposed under various settings. However, most of them only focus on the standard minimization problem, which are incapable of solving the emerging machine learning models. Of particular interest of this paper is the  bilevel optimization problem, which covers numerous machine learning applications, e.g., model-agnostic meta-learning  \cite{finn2017model}, hyperparameter optimization \cite{feurer2019hyperparameter,franceschi2017forward} and neural network architecture search \cite{elsken2019neural,mendoza2016towards,liu2018darts}.  To facilitate federated learning for bilevel optimization problems, in this paper, we aim to develop new optimization algorithms for the \textit{federated  bilevel optimization problem}, which is defined as follows:
\begin{equation} \label{loss_bilevel}
	\begin{aligned}
		& \min_{x} \frac{1}{K}\sum_{k=1}^{K} f^{(k)}(x, y^*(x)) \\
		& s.t. \quad   \ y^*(x) =\arg\min_{y} \frac{1}{K}\sum_{k=1}^{K} g^{(k)}(x, y) \ , 
	\end{aligned}
\end{equation}
where $K$ is the total number of devices, $ f^{(k)}(x, y^*(x)) = \mathbb{E}_{\xi \sim \mathcal{A}^{(k)}}[f^{(k)}(x, y^*(x); \xi) ]$ denotes the loss function of the upper-level optimization problem  on the $k$-th device, $g^{(k)}(x, y)=\mathbb{E}_{\zeta \sim \mathcal{B}^{(k)}}[g^{(k)}(x, y; \zeta) ]$ represents the  loss function of the lower-level optimization problem on the $k$-th device, $\mathcal{A}^{(k)}$ and $\mathcal{B}^{(k)}$ are two data distributions on the $k$-th device.  In this paper, we assume the data distributions across devices are homogeneous.

To solve the stochastic bilevel optimization problem, a wide variety of stochastic gradient based algorithms under the single-machine setting have been developed in the past few years. Specifically, \cite{ghadimi2018approximation} developed a double-loop  stochastic gradient descent algorithm, where the model parameter $y$ of the lower-level optimization problem is updated by stochastic gradient descent (SGD) for multiple iterations in the inner-loop,  and then the model parameter $x$ is updated for one iteration.  The convergence rate of this algorithm is established for nonconvex-strongly-convex problems. Later, the convergence rate is further improved in \cite{hong2020two,ji2021bilevel}  by new algorithmic design, e.g., using a  large mini-batch size. Recently, a couple of single-loop algorithms have been proposed where the model parameter $x$ and $y$ are updated simultaneously. Among them, the momentum-based algorithm has attracted much attention. For instance, \cite{guo2021stochastic} applied the moving-average momentum to the stochastic bilevel optimization problem and established its convergence rate. \cite{yang2021provably,guo2021randomized,khanduri2021momentum} applied the momentum-based variance reduction technique to optimize the stochastic bilevel optimization problem and obtained a better convergence rate. However, all these algorithms just focus on the single-machine setting. When the data is distributed on multiple devices and cannot be shared, new  optimization algorithms should be developed to coordinate the collaboration among multiple devices for learning well-generalizing  machine learning models. 

Federated Learning provides a feasible way to address the aforementioned challenge. Specifically, in federated learning, the machine learning model is shared across multiple devices, rather than the raw data. More specifically, the machine model is updated on each device for multiple iterations and then shared across devices. Based on this learning scheme, numerous federated optimization algorithms \cite{stich2018local,yu2019linear,karimireddy2020scaffold,gao2021convergence} have been developed. For instance, \cite{stich2018local} studied the convergence rate of Local SGD for strongly-convex problems. \cite{yu2019linear} established the convergence rate of momentum Local SGD for nonconvex problems. \cite{gao2021convergence} investigated the compression technique for Local SGD to reduce the communication cost.  However, all these studies only focus on the standard minimization problem. As a result, they cannot be applied to the federated bilevel optimization problem in Eq.~(\ref{loss_bilevel}).   In particular, due to the bilevel structure in Eq.~(\ref{loss_bilevel}), the stochastic hypergradient requires to compute Hessian and Jacobian matrices of the low-level function. It is still unclear if these high-dimensional matrices should be communicated. Moreover, the stochastic hypergradient is NOT an unbiased estimator of the full gradient. As such, traditional federated optimization algorithms and theories do not hold for federated bilevel optimization problems.  
Thus, it is of vital importance to develop new federated optimization algorithms to solve Eq.~(\ref{loss_bilevel}) and provide their theoretical foundations.

\subsection{Contribution}
In this paper, we developed the momentum-based optimization algorithms for  federated  bilevel optimization problems. In particular, we developed a local bilevel stochastic  gradient  with momentum (LocalBSGM) algorithm where each device leverages the momentum stochastic gradient to update the model parameter $x$ and $y$ locally for multiple iterations and then communicates the updated model parameters with the central server.  Moreover, we established the convergence rate of LocalBSGM, demonstrating how the communication period and the number of devices affect the sample and communication complexities. Importantly, our convergence rate indicates that  LocalBSGM  is able to achieve the linear speedup with respect to the number of devices. To the best of our knowledge, this is the first algorithm achieving such a linear speedup result. 

%However, it is challenging to establish its convergence rate due to the interaction between the bilevel structure and the gradient momentum. In this paper, we addressed this challenging problem by a novel theoretical analysis strategy, establishing the convergence rate of our algorithm. To the best of our knowledge, this is the first federated optimization algorithm with theoretical guarantees for Eq.~(\ref{loss_bilevel}). 

Moreover, to further improve the convergence rate, we developed a new local  bilevel stochastic  gradient with momentum-based variance reduction (LocalBSGVR) algorithm,  where each device employs a momentum-based stochastic variance-reduced gradient to update the model parameter $x$ and $y$ simultaneously to reduce the computational cost and accordingly save the communication cost. Compared with LocalBSGM, the variance-reduced gradient estimator makes LocalBSGVR converge faster. In particular, our theoretical analysis demonstrates that LocalBSGVRM is able to achieve better sample and communication complexities than LocalBSGM. Meanwhile, it also enjoys the linear speedup with respect to the number of devices. To the best of our knowledge, this is the first algorithm achieving such sample and communication complexities for federated bilevel optimization problems.  Finally, we conducted extensive experiments and the experimental results confirm the effectiveness of our proposed two algorithms. 
To sum up, we made the following contributions in this paper.
\begin{itemize}
	\item We developed a novel local bilevel stochastic  gradient  with momentum algorithm for the federated bilevel optimization problem and established its convergence rate. 
	\item We proposed a novel local  bilevel stochastic  gradient with momentum-based variance reduction algorithm to improve the convergence rate of our first algorithm, which can achieve better sample and communication complexities. 
\end{itemize}

\begin{table*}[!h] 
	\begin{center}
		\begin{tabular}{l|cc|c}
			%	\hline
			\toprule
			{Methods} &   {Sample Complexity} &  {Communication Complexity}  & Linear Speedup  \\
			\hline
			FEDNEST \cite{tarzanagh2022fednest} & $O(\frac{1}{\epsilon^2})$   &  $O(\frac{1}{\epsilon^2})$  & \xmark\\
			FedBiOAcc \cite{li2022local} & $O(\frac{1}{\epsilon^{3/2}})$   & $O(\frac{1}{\epsilon })$ & \xmark\\
			\hline
			LocalBSGM (Corollary~\ref{corollary_momentum_epsilon}) & $O(\frac{1}{K\epsilon^{2}})$   &  $O(\frac{1}{\epsilon^{3/2}})$ & \cmark\\
			LocalBSGVR (Corollary~\ref{corollary_storm_fixed_epsilon}) & $O(\frac{1}{K\epsilon^{3/2}})$   &  $O(\frac{1}{\epsilon})$ & \cmark\\
			LocalBSGVR (Corollary~\ref{corollary_storm_decay_epsilon}) & $\tilde{O}(\frac{1}{K\epsilon^{3/2}})$   &  $\tilde{O}(\frac{1}{\epsilon})$ & \cmark \\
			\bottomrule 
		\end{tabular}
	\end{center}
	\caption{The  sample and communication complexities of our method and baseline methods to achieve the $\epsilon$-accuracy stationary solution.   $K$ is the number of devices. 
	}
	\label{sequence}
\end{table*}

\section{Related Work}

\subsection{Stochastic Bilevel Optimization}
To solve the bilevel optimization problem in machine learning, a large number of  gradient-based bilevel optimization algorithms have been proposed in recent years. Especially, \cite{ghadimi2018approximation} developed a stochastic gradient based algorithm and established its convergence rate. After that,  a series of algorithms were proposed to improve the convergence rate. For instance, \cite{hong2020two} developed a two-timescale based algorithm to coordinate the updates of the model parameter in the upper-level problem and the lower-level problem. \cite{ji2021bilevel} proposed to use a large batch size to improve the convergence rate. Recently, inspired by the variance reduction technique in the standard minimization problem, a couple of variance-reduced algorithms have been proposed to improve the convergence rate. For instance, \cite{guo2021stochastic} incorporated the momentum technique into  bilevel stochastic gradient descent and established its convergence rate. However, its theoretical convergence rate is in the same order with that  of  \cite{ji2021bilevel}.  \cite{yang2021provably,guo2021randomized,khanduri2021momentum}  incorporated the momentum-based variance reduction technique \cite{cutkosky2019momentum} to bilevel stochastic gradient descent, which orderwisely improves the convergence rate of other algorithms.  In addition, \cite{yang2021provably} employed the SPIDER gradient estimator \cite{fang2018spider} to reduce the gradient variance. As such, as \cite{yang2021provably,guo2021randomized,khanduri2021momentum}, this algorithm also enjoys a better convergence rate than \cite{ji2021bilevel}.
More recently, a couple of gradient-based bilevel optimizers  \cite{huang2021enhanced,huang2021biadam} have been proposed to deal with the nonsmooth problem and the adaptive learning rate.  However, all these algorithms only studied the single-machine setting. Thus, it is of importance to develop federated optimization algorithms for solving Eq.~(\ref{loss_bilevel}).

\subsection{Federated Learning}
With the widespread applications of Federated Learning in computer vision and machine learning, a wide variety of federated optimization algorithms have been proposed in recent years to address different challenges.  For instance, \cite{stich2018local}  established the convergence rate of Local SGD for strongly-convex problems, while \cite{yu2019parallel} established that for nonconvex problems. \cite{li2020federated}  developed FedProx and \cite{karimireddy2020scaffold} proposed SCAFFOLD to address the heterogeneous data distribution issue.  \cite{reisizadeh2020fedpaq,haddadpour2021federated,gao2021convergence} studied how to reduce the communication cost for Local SGD via compressing gradients. \cite{murata2021bias,khanduri2021stem} studied how to reduce the computation cost with advanced stochastic gradient estimators.  However, all these federated learning methods only focus on the traditional minimization problem so that they cannot be applied to federated bilevel optimization problems.

In prepariation for this work, we are aware of two concurrent works. In fact, our work is significantly different from those two works. In particular, \cite{tarzanagh2022fednest} did not leverage the momentum technique so that its convergence rate is inferior to ours. Moreover, it fails to achieve the linear speedup with respect to the number of devices. As for \cite{li2022local}, it studied a much simpler model, where the lower-level function only depends on the local information, rather than the global information as our model. In particular,  the model considered in \cite{li2022local} is defined as follows:
\begin{equation} 
	\begin{aligned}
		& \min_{x} \frac{1}{K}\sum_{k=1}^{K} f^{(k)}(x, y^*(x)^{(k)} ) \\
		& s.t. \quad   \ y^*(x)^{(k)} =\arg\min_{y}  g^{(k)}(x, y) \ , 
	\end{aligned}
\end{equation}
where $ y^*(x)^{(k)}$ is the optimal solution of the local lower-level optimization problem:  $\min_{y}  g^{(k)}(x, y)$, rather than the global one:  $\min_{y}  \frac{1}{K}\sum_{k=1}^{K} g^{(k)}(x, y)$. As a result, the algorithmic design and theoretical analysis are totally different. Moreover, the convergence rate in \cite{li2022local}  cannot achieve the linear speedup regarding the number of devices, while our two algorithms can achieve that. The detailed comparison can be found in Table~\ref{sequence}. All in all, our work is totally different from these two concurrent works.

\section{Preliminaries}

\paragraph{Stochastic Hypergradient.} For Eq.~(\ref{loss_bilevel}), to compute the stochastic hypergradient of the upper-level function regarding the model parameter $x$, we first introduce the auxiliary function $\Phi^{(k)}(x) = f^{(k)}(x, y^*(x)) $ and $\Phi(x) = \frac{1}{K}\sum_{k=1}^{K}\Phi^{(k)}(x)$ where  $y^*(x)$ denotes the optimal solution of the lower-level optimization problem. Then, we can compute the full hypergradient  regarding $x$ on the $k$-th device based on Lemma~\ref{lemma_hypergradient_homo}

\begin{lemma}\label{lemma_hypergradient_homo}
	When the data  distributions across all devices are homogeneous, the hypergradient on the $k$-th device is
\begin{equation}
	\begin{aligned}
		& \nabla \Phi^{(k)}(x)  = \nabla_x f^{(k)}(x, y^*(x)) - \nabla_{xy}^2 g^{(k)}(x, y^*(x)) H\nabla_y f^{(k)}(x, y^*(x)) \ , 
	\end{aligned}
\end{equation}
where $H=[\nabla_{yy}^2g^{(k)}(x, y^*(x))]^{-1}$. 
\end{lemma}

However, this hypergradient is typically infeasible to compute because the optimal solution $y^*(x)$ is expensive to obtain in practice.  Then, we introduce the following $\tilde{\nabla} \Phi^{(k)}(x)$ to approximate it:
\begin{equation}
	\begin{aligned}
		& \tilde{\nabla} \Phi^{(k)}(x)  = \nabla_x f^{(k)}(x, y)- \nabla_{xy}^2 g^{(k)}(x, y) \tilde{H}\nabla_y f^{(k)}(x, y) \ , 
	\end{aligned}
\end{equation}
where $\tilde{H} = [\nabla_{yy}^2g^{(k)}(x, y)]^{-1}$.  It can be observed that this approximator does not require to leverage the optimal solution $y^*(x)$. However, it still needs to compute the inverse of Hessian matrix.  A commonly strategy \cite{yang2021provably} to address this issue  is to use the following approximated gradient:
\begin{equation}
	\begin{aligned}
		&  \hat{\nabla} \Phi^{(k)}(x, y)  = \nabla_x f^{(k)}(x, y)  -  \nabla_{xy}^2g^{(k)}(x, y)\hat{H}\nabla_y f^{(k)}(x, y)  \ , \\
		%		& =  \nabla_x f^{(k)}(x, y) -  \nabla_{xy}^2g^{(k)}(x, y) \Big(\theta\sum_{q=0}^{Q}(I-\theta\nabla_{yy}^2g^{(k)}(x, y))^{q}\Big)\nabla_y f^{(k)}(x, y) \\
	\end{aligned}
\end{equation}
where $\hat{H}=\theta \sum_{q=-1}^{Q-1}\prod_{j=Q-q}^{Q}(I-\theta\nabla_{yy}^2g^{(k)}(x, y))$ and $\theta>0$ and $Q>0$ are hyperparameters.  Based on this approximation, we can compute the stochastic hypergradient regarding $x$ as follows:
\begin{equation}
	\begin{aligned}
		& \hat{\nabla} \Phi^{(k)}(x, y; \hat{\xi})  = \nabla_x f^{(k)}(x, y; \xi)  -  \nabla_{xy}^2g(x, y; \zeta)\hat{H}_{\hat{\xi}}\nabla_y f(x, y; \xi)  \ , \\
	\end{aligned}
\end{equation}
where $\hat{H}_{\hat{\xi}}=\theta \sum_{q=-1}^{Q-1}\prod_{j=Q-q}^{Q}(I-\theta\nabla_{yy}^2g(x, y; \zeta_j))$ and $\hat{\xi}=\{\xi, \zeta_j\}$.  Then, we can leverage this stochastic hypergradient to update the model parameter $x$. 

\paragraph{Assumptions.} To investigate the convergence rate of  federated bilevel optimization algorithms, we introduce the following assumptions, which are commonly used in existing bilevel optimization works \cite{yang2021provably,chen2021closing,khanduri2021momentum}. 
\begin{assumption} \label{assumption_bi_strong}
	For any $k\in \{1,2,\cdots, K\}$, the function $g^{(k)}(x, y; \zeta)$ is $\mu$-strongly convex regarding $y$ for any $x$. 
\end{assumption}

\begin{assumption} \label{assumption_bi_smooth}
	For any $k\in \{1,2,\cdots, K\}$,  the function $f^{(k)}(x, y; \xi)$ and $g^{(k)}(x, y; \zeta)$ satisfy:
	\begin{itemize}
		\item $f^{(k)}(x, y; \xi)$ is $L_0$-Lipschitz continuous.
		\item $\nabla f^{(k)}(x, y; \xi)$ and $\nabla g^{(k)}(x, y; \zeta)$ are $L_1$-Lipschitz continuous.
		\item $\nabla_{xy}^2 g^{(k)}(x, y; \zeta)$ is $L_{21}$-Lipschitz continuous. $\nabla_{yy}^2 g^{(k)}(x, y; \zeta)$ is $L_{22}$-Lipschitz continuous. 
	\end{itemize}
\end{assumption}

\begin{assumption} \label{assumption_bi_var}
	For any $k\in \{1,2,\cdots, K\}$,  the variance of $\nabla g^{(k)}(x, y; \zeta)$ satisfies: $\mathbb{E}[\|\nabla g^{(k)}(x, y) - \nabla g^{(k)}(x, y; \zeta)\|^2]\leq \sigma^2$. 
\end{assumption}

In terms of  \cite{yang2021provably}, we can know that $\Phi^{(k)}(x)$ is $L_{\Phi}$-smooth where $L_{\Phi}=L_1+\frac{2L_1^2+L_{21}L_0^2}{\mu} +\frac{L_{22}L_1L_0+L_1^3+L_{12}L_0L_1}{\mu^2} +\frac{L_{22}L_1^2L_0}{\mu^3}$.  Additionally, we introduce $\hat{L}^2=2L_1^2+4\theta^2L_0^2L_{21}^2(Q+1)^2+8\theta^2L_1^4(Q+1)^2+2\theta^4L_0^2L_1^2L_{22}^2Q^2(Q+1)^2$,  $\tilde{L}=L_1+\frac{L_1^2}{\mu}+\frac{L_0L_{21}}{\mu}+\frac{L_0L_1L_{22}}{\mu^2}$,  $G=2L_0^2+12\theta^2L_0^2L_1^2(Q+1)^2+4\theta^4L_0^2L_1^2(Q+2)(Q+1)^2\sigma^2$, and $\Delta_Q=\frac{(1-\theta\mu)^{Q+1}L_0L_1}{\mu}$.  Their definitions can be found in the  appendix.
Throughout this paper, we denote $\bar{a}_t=\frac{1}{K}\sum_{k=1}^{K}a_t^{(k)}$, where $a^{(k)}$ denotes any local variables on the $k$-th device in the $t$-th iteration.

\section{Momentum-Based Algorithms for Federated Bilevel Optimization Problems}
\begin{algorithm}[]
	\caption{LocalBSGM}
	\label{alg_LocalBSGDM}
	\begin{algorithmic}[1]
		\REQUIRE ${x}_{0}^{(k)}={x}_{0}$, ${y}_{0}^{(k)}={y}_{0}$, $p>1$, $\eta>0$, $\alpha>0$, $\beta>0$, $\rho_1>0$, $\rho_2>0$.\\ Conduct following steps for all devices.
		\FOR{$t=0,\cdots, T-1$} 
		\IF {$t==0$}
		\STATE With the batch size being $1$, compute:
		\STATE $u_{t}^{(k)}  =  \hat{\nabla} \Phi^{(k)}(x_t^{(k)}, y_t^{(k)}; \hat{\xi}_{t}^{(k)} )$,  \\
		$v_{t}^{(k)}  = \nabla_y g^{(k)}(x_t^{(k)}, y_t^{(k)}; \zeta_{t}^{(k)} )$, \\
		\ELSE
		\STATE With the batch size being $1$, compute:
		\STATE $u_{t}^{(k)}  = (1-\alpha\eta)u_{t-1}^{(k)}+\alpha\eta  \hat{\nabla} \Phi^{(k)}(x_t^{(k)}; \hat{\xi}_{t}^{(k)} )$, \\
		\STATE $v_{t}^{(k)}  = (1-\beta\eta)v_{t-1}^{(k)}+\beta \eta\nabla_y g^{(k)}(x_t^{(k)}, y_t^{(k)}; \zeta_{t}^{(k)} )$, \\
		\ENDIF
		%		\STATE $u_{t}^{(k)}  = (1-\alpha)u_{t-1}^{(k)}+\alpha\nabla_x f^{(k)}(x_t^{(k)}, y_t^{(k)}; \xi_{t}^{(k)} )$, \\
		%		\STATE $v_{t}^{(k)}  = (1-\beta)v_{t-1}^{(k)}+\beta \nabla_y f^{(k)}(x_t^{(k)}, y_t^{(k)}; \xi_{t}^{(k)} )$, \\
		\STATE 
		${x}_{t+1}^{(k)}={x}_{t}^{(k)}- \rho_{1}\eta {u}_{t}^{(k)}$, \\
		${y}_{t+1}^{(k)}= {y}_{t}^{(k)}-\rho_{2}\eta{v}_{t}^{(k)}$, \\
		
		\IF {mod($t+1$, $p$)=0}
		\STATE Upload $u_{t+1}^{(k)}$, $v_{t+1}^{(k)}$, $x_{t+1}^{(k)}$, $y_{t+1}^{(k)}$ to the central server and reset local variables with the global one: \\
		$u_{t+1}^{(k)} = \bar{u}_{t+1}=\frac{1}{K}\sum_{k'=1}^{K}u_{t+1}^{(k')} $  ,  \\ 
		$v_{t+1}^{(k)} = \bar{v}_{t+1}=\frac{1}{K}\sum_{k'=1}^{K}v_{t+1}^{(k')} $  ,  \\
		$x_{t+1}^{(k)} = \bar{x}_{t+1}=\frac{1}{K}\sum_{k'=1}^{K}x_{t+1}^{(k')} $ ,\\ 
		$y_{t+1}^{(k)} = \bar{y}_{t+1}=\frac{1}{K}\sum_{k'=1}^{K}y_{t+1}^{(k')} $ ,
		
		\ENDIF

		\ENDFOR
	\end{algorithmic}
\end{algorithm}

\subsection{Local Bilevel Stochastic Gradient Descent with Momentum}
In this subsection, we develop a novel local bilevel stochastic gradient descent with momentum algorithm in Algorithm~\ref{alg_LocalBSGDM}. In detail,  each device $k$ computes the momentum for  stochastic (hyper-)gradients $\hat{\nabla} \Phi^{(k)}(x_t^{(k)}, y_t^{(k)}; \hat{\xi}_{t}^{(k)} )$ and $\nabla_y g^{(k)}(x_t^{(k)}, y_t^{(k)}; \zeta_{t}^{(k)} )$ as follows:
\begin{equation}
	\begin{aligned}
		& u_{t}^{(k)}  = (1-\alpha\eta)u_{t-1}^{(k)}+\alpha\eta  \hat{\nabla} \Phi^{(k)}(x_t^{(k)},y_t^{(k)}; \hat{\xi}_{t}^{(k)} )  \ , \\
		& v_{t}^{(k)}  = (1-\beta\eta)v_{t-1}^{(k)}+\beta \eta\nabla_y g^{(k)}(x_t^{(k)}, y_t^{(k)}; \zeta_{t}^{(k)} )  \ , \\
	\end{aligned}
\end{equation}
where $\alpha>0$, $\beta>0$, $\eta>0$ are hyperparameters,  and $\alpha\eta<1$, $\beta\eta<1$. It can be observed that $u_{t}^{(k)}$ and $v_{t}^{(k)}$ are the moving-average estimation for those two stochastic gradients. Then, the $k$-th device leverages those two momentum to update its local model parameters as follows:
\begin{equation}
	\begin{aligned}
		& {x}_{t+1}^{(k)}={x}_{t}^{(k)}- \rho_{1}\eta {u}_{t}^{(k)} \ , \\
		& {y}_{t+1}^{(k)}= {y}_{t}^{(k)}-\rho_{2}\eta{v}_{t}^{(k)}  \  , \\
	\end{aligned}
\end{equation}
where $\rho_1>0$ and $\rho_2>0$ are two hyperparameters.  As the standard federated optimization algorithm, each device uploads both local momentum and model parameters to the central server at every $p$ (where $p>1$) iterations. Then,  local model parameters and momentum are reset to the global one,  which is shown in Line 12 of Algorithm~\ref{alg_LocalBSGDM}.

In Theorem~\ref{theorem_bi_sgdm}, we establish the convergence rate of Algorithm~\ref{alg_LocalBSGDM}. 
\begin{theorem} \label{theorem_bi_sgdm}
	Suppose Assumptions~\ref{assumption_bi_strong}-\ref{assumption_bi_var} hold, by setting $\alpha>0$, $\beta>0$, $\theta<\frac{1}{L_1}$, $\rho_1\leq \Big\{\frac{3\rho_{2}\mu^2}{50L_1\tilde{L}} , \frac{1}{4} \Big/\sqrt{\frac{4 \hat{L}^2}{\alpha^2} + \frac{500 L_1^2\tilde{L}^2}{3\beta^2\mu^2} }\Big\}$, and $\rho_2\leq \Big\{ \frac{6\tilde{L}^2}{\mu}  \Big/ \Big(\frac{4\hat{L}^2}{\alpha^2} +\frac{400 L_1^2\tilde{L}^2}{3\beta^2\mu^2} \Big), \frac{1}{6L_1}\Big\} $, $\eta<\min\{\frac{1}{\alpha}, \frac{1}{\beta}, \frac{1}{2\rho_{1}L_{\Phi}}, \frac{1}{3p(\alpha^2+\beta^2)^{1/4} (\rho_{1}^2+\rho_{2}^2)^{1/4}\hat{L}^{1/2} } , 1\}$,  Algorithm~\ref{alg_LocalBSGDM} has the following convergence rate: 
	\begin{equation}
		\begin{aligned}
			&  \quad \frac{1}{T}\sum_{t=0}^{T-1}\Big(\mathbb{E}[\|\nabla \Phi(\bar{  {x}}_{t})\|^2]+ \tilde{L}^2\mathbb{E}[\|\bar{y}_t - y^*(\bar{x}_t)\|^2]\Big) \leq \frac{2( \Phi(x_{0}) - \Phi(x_{*}) )}{\eta\rho_1T} +  \frac{20\tilde{L}^2}{\eta T \rho_{2}\mu } \|\bar{   {y}}_{0} -    {y}^{*}(\bar{   {x}}_{0})\| ^2 \\
			& \quad + \frac{4G^2}{\alpha\eta T}  +  \frac{500 \tilde{L}^2\sigma^2 }{3\eta T\beta\mu^2}  + \frac{4\alpha\eta\sigma^2}{ K}  + \frac{500 \beta\eta\sigma^2\tilde{L}^2}{3K\mu^2}+  4\Delta_Q^2\\
			&  \quad + 48\alpha^2\eta^4p^4(\rho_{1}^2+\rho_{2}^2)\Big(2\hat{L}^2+  \frac{500L_1^2\tilde{L}^2}{3\mu^2}  \Big)G^2 +48\beta^2p^4\eta^4(\rho_{1}^2+\rho_{2}^2)\Big(2\hat{L}^2+  \frac{500L_1^2\tilde{L}^2}{3\mu^2}  \Big) \sigma^2\\
			& \quad+48\alpha^2p^2\eta^2(\rho_{1}^2 +\rho_{2}^2)\Big(\frac{4 \hat{L}^2}{\alpha^2} +\frac{500 L_1^2 \tilde{L}^2}{3\beta^2\mu^2} \Big) G^2 +48\beta^2p^2\eta^2(\rho_{1}^2 +\rho_{2}^2)\Big(\frac{4 \hat{L}^2}{\alpha^2} +\frac{500 L_1^2 \tilde{L}^2}{3\beta^2\mu^2} \Big) \sigma^2 \ .\\
		\end{aligned}
	\end{equation}
\end{theorem}
\begin{corollary} \label{corollary_momentum_t}
	Suppose Assumptions~\ref{assumption_bi_strong}-\ref{assumption_bi_var} hold, by setting $p=O(\frac{T^{1/4}}{K^{3/4}})$,  $\eta=O(\frac{K^{1/2}}{T^{1/2}})$, and $Q=O(\log(K^{1/2}T^{1/2}))$,  and other hyperparameters $\alpha$, $\beta$, $\rho_1$, $\rho_2$  to be independent of $T$ and $K$,  Algorithm~\ref{alg_LocalBSGDM} has the following convergence rate: 
	\begin{equation}
		\begin{aligned}
			&  \frac{1}{T}\sum_{t=0}^{T-1}\Big(\mathbb{E}[\|\nabla \Phi(\bar{  {x}}_{t})\|^2]+ \tilde{L}^2\mathbb{E}[\|\bar{y}_t - y^*(\bar{x}_t)\|^2]\Big) \leq O(\frac{1}{\sqrt{KT}})  \ . \\
		\end{aligned}
	\end{equation} 
\end{corollary}

\begin{corollary} \label{corollary_momentum_epsilon}
	Suppose Assumptions~\ref{assumption_bi_strong}-\ref{assumption_bi_var} hold, by setting  $T=O(\frac{1}{K\epsilon^2})$,  $p=O(\frac{1}{K\epsilon^{1/2}})$,  $\eta=O(K\epsilon)$, and $Q=O(\log(\frac{1}{\epsilon}))$, and other hyperparameters $\alpha$, $\beta$, $\rho_1$, $\rho_2$  to be independent of $\epsilon$ and $K$,  Algorithm~\ref{alg_LocalBSGDM} can achieve the $\epsilon$-accuracy stationary point: 
	\begin{equation}
		\begin{aligned}
			&   \frac{1}{T}\sum_{t=0}^{T-1}\Big(\mathbb{E}[\|\nabla \Phi(\bar{  {x}}_{t})\|^2]+ \tilde{L}^2\mathbb{E}[\|\bar{y}_t - y^*(\bar{x}_t)\|^2]\Big) \leq \epsilon \ . \\
		\end{aligned}
	\end{equation} 
\end{corollary}

\begin{remark}
	From Corollary~\ref{corollary_momentum_epsilon}, we can  know that the iteration complexity (i.e., sample complexity) is  $T=O(\frac{1}{K\epsilon^2})$, which indicates that our LocalBSGM can achieve linear speedup with respect to the number of devices.   Moreover, the communication complexity of our algorithm is $T/p=O(\frac{1}{\epsilon^{3/2}})$.   On the contrary, FEDNEST \cite{tarzanagh2022fednest} can only achieve $O(\frac{1}{\epsilon^2})$ sample and communication complexities, which are inferior to ours. 
	
\end{remark}

\begin{algorithm}[ht]
	\caption{LocalBSGVRM}
	\label{alg_LocalBSGDVRM}
	\begin{algorithmic}[1]
		\REQUIRE ${x}_{1}^{(k)}={x}_{1}$, ${y}_{1}^{(k)}={y}^*(x_1)$, $p>1$,  $\alpha>0$, $\beta>0$, $\rho_1>0$, $\rho_2>0$.\\ Conduct following steps for all devices.
		\FOR{$t=0,\cdots, T-1$} 
		\IF {$t==0$}
		\STATE With the batch size being $B>1$, compute:
		\STATE $u_{t}^{(k)}  =  \hat{\nabla} \Phi^{(k)}(x_t^{(k)}, {y}_{t}^{(k)}; \hat{\xi}_{t}^{(k)} )$, \\
		$v_{t}^{(k)}  = \nabla_y g^{(k)}(x_t^{(k)}, y_t^{(k)}; \zeta_{t}^{(k)} )$, \\
		\ELSE
		\STATE With the batch size being $1$, compute:
		\STATE $u_{t}^{(k)}  = (1-\alpha\eta_{t-1}^2)(u_{t-1}^{(k)}- \hat{\nabla} \Phi^{(k)}(x_{t-1}^{(k)}, y_{t-1}^{(k)}; \hat{\xi}_{t}^{(k)} ))+ \hat{\nabla} \Phi^{(k)}(x_t^{(k)}, y_{t}^{(k)}; \hat{\xi}_{t}^{(k)} )$, \\
		\STATE $v_{t}^{(k)}  = (1-\beta\eta_{t-1}^2)(v_{t-1}^{(k)}-\nabla_y g^{(k)}(x_{t-1}^{(k)}, y_{t-1}^{(k)}; \zeta_{t}^{(k)} ))+ \nabla_y g^{(k)}(x_t^{(k)}, y_t^{(k)}; \zeta_{t}^{(k)} )$, \\
		\ENDIF
		%		\STATE $u_{t}^{(k)}  = (1-\alpha)u_{t-1}^{(k)}+\alpha\nabla_x f^{(k)}(x_t^{(k)}, y_t^{(k)}; \xi_{t}^{(k)} )$, \\
		%		\STATE $v_{t}^{(k)}  = (1-\beta)v_{t-1}^{(k)}+\beta \nabla_y f^{(k)}(x_t^{(k)}, y_t^{(k)}; \xi_{t}^{(k)} )$, \\
		\STATE 
		${x}_{t+1}^{(k)}={x}_{t}^{(k)}- \rho_{1}\eta_t {u}_{t}^{(k)}$, \\
		${y}_{t+1}^{(k)}= {y}_{t}^{(k)}-\rho_{2}\eta_t{v}_{t}^{(k)}$ \\
		
		\IF {mod($t+1$, $p$)=0}
		\STATE Upload $u_{t+1}^{(k)}$, $v_{t+1}^{(k)}$, $x_{t+1}^{(k)}$, $y_{t+1}^{(k)}$ to server and reset \\
		$u_{t+1}^{(k)} = \bar{u}_{t+1}=\frac{1}{K}\sum_{k'=1}^{K}u_{t+1}^{(k')} $  ,\\ 
		$v_{t+1}^{(k)} = \bar{v}_{t+1}=\frac{1}{K}\sum_{k'=1}^{K}v_{t+1}^{(k')} $  , \\
		$x_{t+1}^{(k)} = \bar{x}_{t+1}=\frac{1}{K}\sum_{k'=1}^{K}x_{t+1}^{(k')} $ ,\\
		$y_{t+1}^{(k)} = \bar{y}_{t+1}=\frac{1}{K}\sum_{k'=1}^{K}y_{t+1}^{(k')} $ ,
		
		\ENDIF

		\ENDFOR
	\end{algorithmic}
\end{algorithm}

\vspace{-10pt}
\subsection{Local Bilevel Stochastic Gradient Descent with Momentum-Based Variance Reduction}

In Algorithm~\ref{alg_LocalBSGDVRM}, we further developed a local bilevel stochastic gradient descent with momentum-based variance reduction (LocalBSGVRM) algorithm. Compared with Algorithm~\ref{alg_LocalBSGDM},  LocalBSGVRM employs a variance-reduced gradient estimator, which was first proposed in \cite{cutkosky2019momentum} for the standard minimization problem,  to accelerate the convergence rate. Specifically, the $k$-th device computes the momentum-based variance-reduced gradient as follows:
\begin{equation}
	\begin{aligned}
		& u_{t}^{(k)}  = (1-\alpha\eta_{t-1}^2)(u_{t-1}^{(k)}- \hat{\nabla} \Phi^{(k)}(x_{t-1}^{(k)}, y_{t-1}^{(k)}; \hat{\xi}_{t}^{(k)} )) + \hat{\nabla} \Phi^{(k)}(x_t^{(k)}, y_{t}^{(k)}; \hat{\xi}_{t}^{(k)} ) \ , \\
		& v_{t}^{(k)}  = (1-\beta\eta_{t-1}^2)(v_{t-1}^{(k)}-\nabla_y g^{(k)}(x_{t-1}^{(k)}, y_{t-1}^{(k)}; \zeta_{t}^{(k)} )) + \nabla_y g^{(k)}(x_t^{(k)}, y_t^{(k)}; \zeta_{t}^{(k)} ) \ , \\
	\end{aligned}
\end{equation}
where $\alpha\eta_t^2<1$ and $\beta\eta_t^2<1$.  With this new gradient estimator, LocalBSGVRM updates and communicates local model parameters in the same way as LocalBSGM.

In Theorem~\ref{theorem_LocalBSGDVRM_constant}, we established the convergence rate of  LocalBSGVRM when the learning rate is fixed.
\begin{theorem} \label{theorem_LocalBSGDVRM_constant}
	Suppose Assumptions~\ref{assumption_bi_strong}-\ref{assumption_bi_var} hold, by setting $\theta<\frac{1}{L_1}$,  $\alpha\leq \frac{\hat{L}^2}{K}$,  $\beta\leq \frac{\hat{L}^2}{K}$,  $\eta_t\equiv \eta\leq \min\Big\{\frac{1}{\sqrt{\alpha}}, \frac{1}{\sqrt{\beta}},  \frac{1}{200p \hat{L}}\Big\}$, and 
	\begin{equation}
		\begin{aligned}
			& \rho_1\leq \min\Bigg\{\frac{\rho_{2}\mu^2} {20\tilde{L}L_1 }, \frac{1}{2}\Bigg/\sqrt{\frac{16\hat{L}^2}{\alpha K} + \frac{2000L_1^2\tilde{L}^2}{3\beta \mu^2 K} + \frac{12}{25}\Big(4  + \frac{1000\tilde{L}^2}{3\mu^2} +\frac{16\hat{L}^2}{\alpha K }  +\frac{2000 L_1^2\tilde{L}^2}{3\beta \mu^2 K}\Big)} , 10\Bigg\} \ , \\
			& \rho_2 \leq \min\Bigg\{\frac{1}{6L_1}, \frac{15\tilde{L}^2}{\mu} \Bigg/\Bigg(\frac{16\hat{L}^2}{\alpha K} + \frac{2000L_1^2\tilde{L}^2}{3\beta \mu^2 K} + \frac{12}{25}\Big(4  + \frac{1000\tilde{L}^2}{3\mu^2} +\frac{16\hat{L}^2}{\alpha K }  +\frac{2000 L_1^2\tilde{L}^2}{3\beta \mu^2 K}\Big)\Bigg) , 10\Bigg\}\ , \\
		\end{aligned}
	\end{equation}
	 Algorithm~\ref{alg_LocalBSGDVRM} has the following convergence rate: 
	\begin{equation}
		\begin{aligned}
			& \quad \frac{1}{T}\sum_{t=0}^{T-1}\Big(\mathbb{E}[\|\nabla \Phi(\bar{  {x}}_{t})\|^2]+ \tilde{L}^2\mathbb{E}[\|\bar{y}_t - y^*(\bar{x}_t)\|^2]\Big)   \\
			& \leq \frac{2(\Phi(x_0)-\Phi(x_*))}{\eta \rho_1T} + \frac{10\tilde{L}^2}{\rho_{2}\mu\eta T}\|\bar{   {y}}_{0} -    {y}^{*}(\bar{   {x}}_{0})\| ^2 +\frac{2G^2}{\alpha\eta^2TKB}   +   \frac{250\tilde{L}^2\sigma^2}{3\beta \mu^2\eta^2 TKB} + 4\Delta_Q^2+  \frac{8\alpha\eta^2G^2}{ K}  \\
			& \quad + \frac{1000\beta\eta^2\sigma^2\tilde{L}^2}{3 \mu^2 K}+(\rho_1^2+\rho_{2}^2)\Big(4  + \frac{1000\tilde{L}^2}{3\mu^2} +\frac{16\hat{L}^2}{\alpha K}  +\frac{2000 L_1^2\tilde{L}^2}{3\beta \mu^2 K}\Big) \times (48\beta^2p^2\eta^4\sigma^2 +48\alpha^2p^2\eta^4 G^2 )  \  .   \\
		\end{aligned}
	\end{equation}
\end{theorem}
\begin{corollary} \label{corollary_storm_fixed_t}
	Suppose Assumptions~\ref{assumption_bi_strong}-\ref{assumption_bi_var} hold, by setting $\alpha=O(\frac{1}{K})$, $\beta=O(\frac{1}{K})$,  $\eta=O(\frac{K^{2/3}}{T^{1/3}})$,  $B=O(\frac{T^{1/3}}{K^{2/3}})$,  $p=O(\frac{T^{1/3}}{K^{2/3}})$,   $Q=O(\log(K^{2/3}T^{2/3}))$,  and $\rho_1$, $\rho_2$ to be independent of $T$ and $K$,   Algorithm~\ref{alg_LocalBSGDVRM} has the following convergence rate: 
	\begin{equation}
		\begin{aligned}
			&   \frac{1}{T}\sum_{t=0}^{T-1}\Big(\mathbb{E}[\|\nabla \Phi(\bar{  {x}}_{t})\|^2]+ \tilde{L}^2\mathbb{E}[\|\bar{y}_t - y^*(\bar{x}_t)\|^2]\Big) \leq O\Big(\frac{1}{K^{2/3}T^{2/3}}\Big)  \ . \\
		\end{aligned}
	\end{equation} 
\end{corollary}

\begin{corollary} \label{corollary_storm_fixed_epsilon}
	Suppose Assumptions~\ref{assumption_bi_strong}-\ref{assumption_bi_var} hold, by setting $\alpha=O(\frac{1}{K})$, $\beta=O(\frac{1}{K})$, $T=O(\frac{1}{K\epsilon^{3/2}})$,  $p=O(\frac{1}{K\epsilon^{1/2} })$,  $\eta=O(K\epsilon^{1/2})$, $B=O(\frac{1}{K\epsilon^{1/2}})$,  $Q=O(\log(\frac{1}{\epsilon}))$, and $\rho_1$, $\rho_2$ to be independent of $\epsilon$ and $K$,  Algorithm~\ref{alg_LocalBSGDVRM} can achieve the $\epsilon$-accuracy stationary point: 
	\begin{equation}
		\begin{aligned}
			&   \frac{1}{T}\sum_{t=0}^{T-1}\Big(\mathbb{E}[\|\nabla \Phi(\bar{  {x}}_{t})\|^2]+ \tilde{L}^2\mathbb{E}[\|\bar{y}_t - y^*(\bar{x}_t)\|^2]\Big) \leq \epsilon \ . \\
		\end{aligned}
	\end{equation} 
\end{corollary}
\begin{remark}
	From Corollary~\ref{corollary_storm_fixed_epsilon}, we can  know that the iteration complexity is  $O(\frac{1}{K\epsilon^{3/2}})$, which is much better than that of our  Algorithm~\ref{alg_LocalBSGDM}. Furthermore, the communication complexity is $T/p=O(\frac{1}{\epsilon})$, which is also much better than $O(\frac{1}{\epsilon^{3/2}})$ of Algorithm~\ref{alg_LocalBSGDM}. On the contrary, even though FedBiOAcc \cite{li2022local} also leverages the variance-reduced gradient estimator, its sample complexity  $O(\frac{1}{\epsilon^{3/2}})$ does not show the linear speedup regarding the number of devices. 
	
\end{remark}

In Theorem~\ref{theorem_LocalBSGDVRM}, we established the convergence rate of  LocalBSGVRM when it has a decaying  learning rate. 
\begin{theorem} \label{theorem_LocalBSGDVRM}
	Suppose Assumptions~\ref{assumption_bi_strong}-\ref{assumption_bi_var} hold, by setting $ \alpha = \frac{\hat{L}^2}{3pK^2}+\frac{\hat{L}^2}{K}$, $\beta = \frac{\hat{L}^2}{3pK^2}+\frac{\hat{L}^2}{K}$,  $\eta_t = \frac{K^{2/3}/\hat{L}}{(w_t+t)^{1/3}} $, $w_t = \max\{2, 200^3K^2p^3 - t, \frac{8\rho_1^3 L_{\Phi}^3K^2}{\hat{L}^3} - t\}$, 
	$\rho_{1}  \leq \min\Big\{\frac{\rho_{2}\mu^2}{60L_1^2}, \frac{\mu}{100\tilde{L}}, 10 \Big\}$,  $\rho_2\leq \Big\{\frac{15\mu}{1174}  , \frac{1}{6L_1}, 10\Big\}$,  $\theta<\frac{1}{L_1}$,  Algorithm~\ref{alg_LocalBSGDVRM} has the following convergence rate: 
	\begin{equation}
		\begin{aligned}
			&\quad  \frac{1}{T}\sum_{t=0}^{T-1} \Big(\mathbb{E}[\|\nabla \Phi(\bar{  {x}}_{t})\|^2 ]   + \tilde{L}^2\mathbb{E}[\|\bar{y}_t - y^*(\bar{x}_t)\|^2]\Big) \\
			& \leq  \frac{2(\Phi(x_0)- \Phi(x_*))}{\rho_1} (\frac{200p\hat{L}}{ T} + \frac{\hat{L}}{K^{2/3} T^{2/3}} ) + (\frac{200p\hat{L}}{ T} + \frac{\hat{L}}{K^{2/3} T^{2/3}} )\frac{20 \tilde{L}^2}{\rho_{2}\mu} \|\bar{   {y}}_{0} -    {y}^{*}(\bar{   {x}}_{0})\| ^2  \\
			& \quad +(\frac{200p}{ T} + \frac{1}{K^{2/3} T^{2/3}} )\frac{800  pG^2 }{B}  + (\frac{200p}{ T} + \frac{1}{K^{2/3} T^{2/3}} )\frac{200000p\sigma^2\tilde{L}^2}{3\mu^2 B}  + (\frac{200p}{ T} + \frac{1}{K^{2/3} T^{2/3}} )32\sigma^2 \log(T+1) \\
			& \quad + (\frac{200p}{ T} + \frac{1}{K^{2/3} T^{2/3}} )\frac{4000  \sigma^2\tilde{L}^2}{3\mu^2}\log(T+1) + ( \frac{4\times200^3K^2p^3}{T} + 4 ) \Delta_Q^2\log (T+1)\\
			& \quad  + (\frac{200p}{ T} + \frac{1}{K^{2/3} T^{2/3}} )\frac{3(\rho_1^2+\rho_2^2)\tilde{L}^2(\sigma^2 +G^2)}{\mu^2} \log(T+1)\  . \\
		\end{aligned}
	\end{equation}
\end{theorem}
\begin{corollary} \label{corollary_storm_decay_t}
	Suppose Assumptions~\ref{assumption_bi_strong}-\ref{assumption_bi_var} hold, by setting $\alpha=O(\frac{1}{K})$, $\beta=O(\frac{1}{K})$,  $\eta=O(\frac{K^{2/3}}{T^{1/3}})$,  $B=O(\frac{T^{1/3}}{K^{2/3}})$,   $p=O(\frac{T^{1/3}}{K^{2/3}})$,   $Q=O(\log(K^{2/3}T^{2/3}))$,  and $\rho_1$, $\rho_2$ to be independent of $T$ and $K$,   Algorithm~\ref{alg_LocalBSGDVRM} has the following convergence rate: 
	\begin{equation}
		\begin{aligned}
			&  \frac{1}{T}\sum_{t=0}^{T-1}\Big(\mathbb{E}[\|\nabla \Phi(\bar{  {x}}_{t})\|^2]+ \tilde{L}^2\mathbb{E}[\|\bar{y}_t - y^*(\bar{x}_t)\|^2]\Big)  \leq {O}\Big(\frac{\log (T+1)}{K^{2/3}T^{2/3}}\Big) \ . \\
		\end{aligned}
	\end{equation} 
\end{corollary}
\begin{remark}
	Compared with the fixed learning rate in Theorem~\ref{theorem_LocalBSGDVRM_constant}, the convergence rate has an additional factor $\log (T+1)$, when using the decayed learning rate 
\end{remark}

\begin{corollary} \label{corollary_storm_decay_epsilon}
	Suppose Assumptions~\ref{assumption_bi_strong}-\ref{assumption_bi_var} hold, by setting $\alpha=O(\frac{1}{K})$, $\beta=O(\frac{1}{K})$, $T=\tilde{O}(\frac{1}{K\epsilon^{3/2}})$,  $p=\tilde{O}(\frac{1}{K\epsilon^{1/2} })$,  $B=\tilde{O}(\frac{1}{\epsilon^{1/2}})$,   $Q=\tilde{O}(\log(\frac{1}{\epsilon}))$,  and $\rho_1$, $\rho_2$ to be independent of $T$ and $K$,   Algorithm~\ref{alg_LocalBSGDVRM} can achieve the $\epsilon$-accuracy stationary point: 
	\begin{equation}
		\begin{aligned}
			&   \frac{1}{T}\sum_{t=0}^{T-1}\Big(\mathbb{E}[\|\nabla \Phi(\bar{  {x}}_{t})\|^2]+ \tilde{L}^2\mathbb{E}[\|\bar{y}_t - y^*(\bar{x}_t)\|^2]\Big)  \leq \epsilon \ . \\
		\end{aligned}
	\end{equation} 
\end{corollary}
\begin{remark}
	From Corollary~\ref{corollary_storm_decay_epsilon}, it is easy to   know that the iteration complexity is  $\tilde{O}(\frac{1}{K\epsilon^{3/2}})$, indicating the linear speedup with respect to the number of devices.  Furthermore, the communication complexity is $T/p=\tilde{O}(\frac{1}{\epsilon})$.
\end{remark}

In summary, we established the convergence rate of our proposed LocalBSGM and LocalBSGVRM. Both of them achieve linear speedup regarding the number of devices.  Additionally, we provided the communication complexity of our two algorithms.  To the best of our knowledge, this is the first work achieving such theoretical results for federated bilevel optimization algorithms.

\section{Conclusion}
In this paper, we developed two novel momentum-based algorithms for the federated stochastic bilevel optimization problems. To the best of our knowledge, this is the first work studying this kind of problem. More importantly, we established the convergence rate of our two algorithms, which enjoy superior sample and communication complexities.

\newpage
%%-----------------------------
\bibliographystyle{abbrv}
\bibliography{egbib}
%%-----------------------------

%\onecolumn
%\appendix
% \newpage
%\input{supp_1}
%% \newpage
%\input{supp_2}
%% \newpage
%\input{supp_3}

\end{document}